%% file: main.tex
\documentclass[sigconf]{acmart}
\usepackage{balance}
\AtBeginDocument{%
}

\usepackage{multirow}

\setcopyright{none}
\renewcommand\footnotetextcopyrightpermission[1]{}
\usepackage{algorithm}
\usepackage{algpseudocode}

\usepackage{mathtools}

\usepackage{xcolor}
\definecolor{bestred}{RGB}{220,20,60}
\definecolor{secondblue}{RGB}{30,90,200}

\begin{document}

\title{Depth-Guided Video Object Counting in Crowded Scenes}

\thispagestyle{plain}
\pagestyle{plain}

\author{Yuanjing Xu}
\email{xuyuanjing@stu.hit.edu.cn}
\affiliation{%
  \institution{Harbin Institute of Technology (Weihai)}
  \city{Weihai}
  \state{Shandong}
  \country{China}
}

\author{Xinyan Liu}
\email{xinyliu@hit.edu.cn}
\authornote{Corresponding author.}
\affiliation{%
  \institution{Harbin Institute of Technology (Weihai)}
  \city{Weihai}
  \state{Shandong}
  \country{China}
}
\affiliation{%
  \institution{City University of Hong Kong}
  \city{Hong Kong}
  \country{China}
}

\author{Weidong Chen}
\email{chenweidong@ustc.edu.cn}
\affiliation{%
  \institution{University of Science and Technology of China}
  \city{Hefei}
  \state{Anhui}
  \country{China}
}

\author{Zixuan Zou}
\email{zouzixuan@stu.hit.edu.cn}
\affiliation{%
  \institution{Harbin Institute of Technology (Weihai)}
  \city{Weihai}
  \state{Shandong}
  \country{China}
}

\author{Linhao Zhang}
\email{zhanglinhao@stu.hit.edu.cn}
\affiliation{%
  \institution{Harbin Institute of Technology (Weihai)}
  \city{Weihai}
  \state{Shandong}
  \country{China}
}

\author{Zhuangzhe Meng}
\email{mengzhuangzhe@stu.hit.edu.cn}
\affiliation{%
  \institution{Harbin Institute of Technology (Weihai)}
  \city{Weihai}
  \state{Shandong}
  \country{China}
}

\author{Antoni B. Chan}
\email{abchan@cityu.edu.hk}
\affiliation{%
  \institution{City University of Hong Kong}
  \city{Hong Kong}
  \country{China}
}

\author{Weigang Zhang}
\email{wgzhang@hit.edu.cn}
\affiliation{%
  \institution{Harbin Institute of Technology (Weihai)}
  \city{Weihai}
  \state{Shandong}
  \country{China}
}
\affiliation{%
  \institution{Harbin Institute of Technology Qingdao Research Institute}
  \city{Qingdao}
  \state{Shandong}
  \country{China}
}
\renewcommand{\shortauthors}{Xu et al.}

\begin{abstract}
Our primary objective is to advance video object counting in crowded scenes, aiming to robustly count all instances of a target category based on given text or visual prompts.
Existing methods rely on RGB information, limiting their discriminative ability in crowded and occluded conditions. 
To address this, we propose a Depth-Guided Detector (DG-Det) along with a general post-processing pipeline. By integrating depth cues with multi-scale RGB-D cross-attention and explicit occlusion prediction, our method enhances spatial understanding and achieves robust detection in crowded and occluded scenes. Furthermore, we introduce a unified de-duplication framework to eliminate cross-frame redundant counting. To facilitate future research, we also release a new RGB-D Video Object Counting dataset featuring depth information and multiple object categories persequence. Extensive experiments demonstrate that our method achieves a 62.01\% reduction in MAE compared to existing baselines, and also produces consistent improvements in RMSE. We provide the source code at \url{https://github.com/streamer-AP/DG-Net} and the dataset at \url{https://huggingface.co/datasets/aerospace123/RGBD-VideoCount.}
\end{abstract}

\ccsdesc[500]{Computing methodologies~Object detection}

\keywords{Object Counting in videos, RGB-D, Object Tracking}

\maketitle

\section{Introduction}
\begin{figure}[t]
    \centering
    \includegraphics[width=\linewidth]{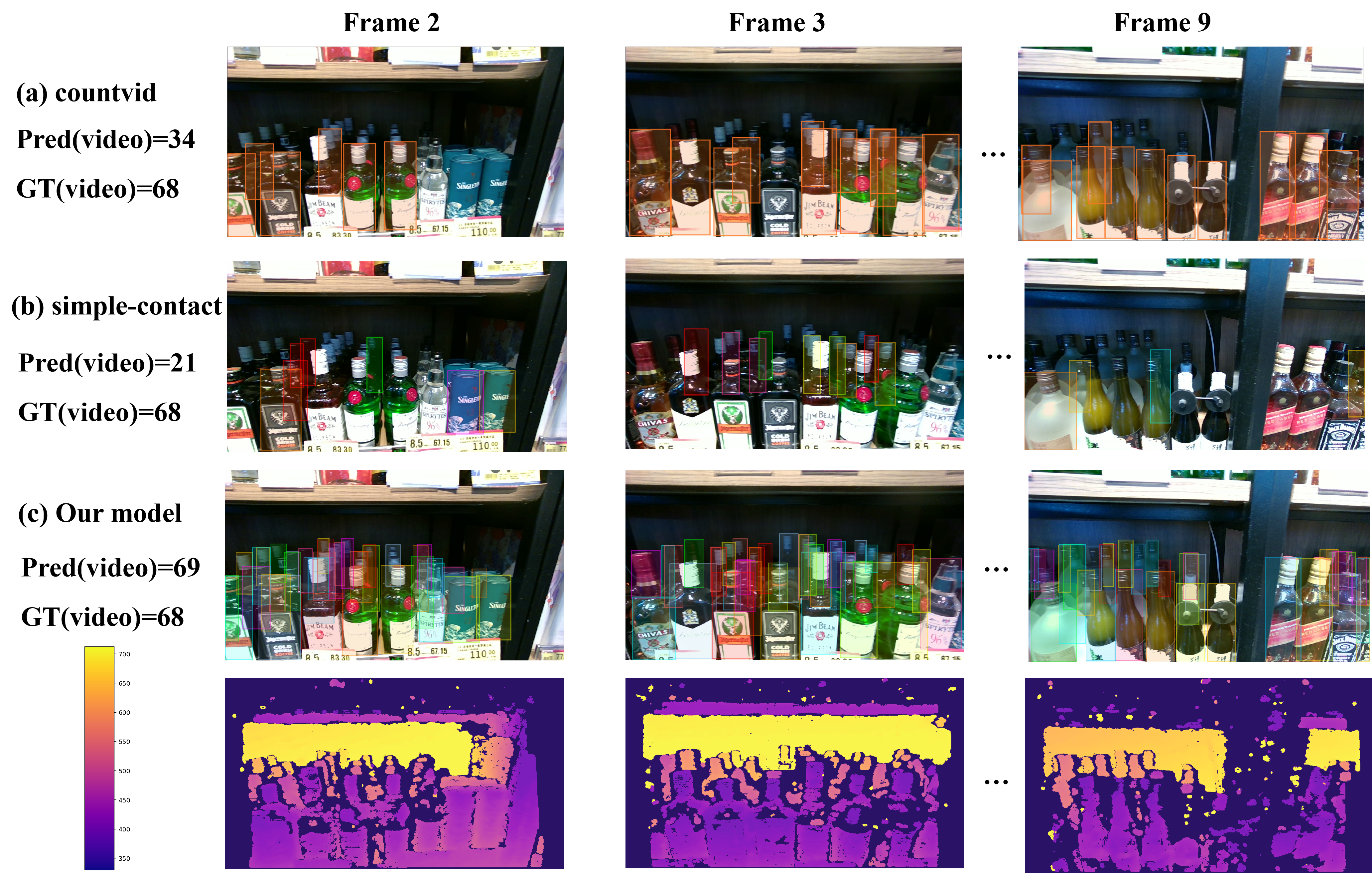}
    \Description{Qualitative comparison of object counting in a dense shelf scene. CountVID misses occluded objects, naive RGB-D concatenation struggles with overlapping targets, and the proposed method uses depth cues to produce more accurate video-level counts.}
    \caption{Qualitative comparison in a dense shelf scene. (a) CountVID misses occluded objects and undercounts. (b) Naive RGB-D concatenation still struggles with overlapping targets. (c) Our method exploits depth cues to better resolve occlusion and achieve more accurate video-level counting.}
    \label{fig:intro}
\end{figure}

We focus on the task of video object counting in crowded scenes, where the goal is to enumerate all instances of a specified target category within a video using either text or visual prompts. However, research on video counting in crowded and occluded scenes remains very limited. In real-world applications such as shelf inventory checks and warehouse inspections, accurate counting from dynamic video streams is a key prerequisite for automated inventory management and out-of-stock warnings, which have substantial practical value. Motivated by this, we focus on video counting in real crowded and occluded object scenes, and conduct the first research on video individual counting in crowded scenes. Unlike common targets in existing video tasks, such as pedestrians and vehicles, same-category objects in multi-category occluded scenes are often highly similar in packaging, texture, and scale. At the same time, these scenes involve dense arrangements, frequent occlusions, and viewpoint changes. Since RGB mainly captures appearance cues such as color and texture, relying solely on RGB features within a single frame makes it difficult to effectively separate tightly adjacent and mutually occluded objects. Across frames, viewpoint variation and short-term occlusion further disturb the stability of appearance representations, making instance matching more prone to confusion. As a result, existing detection-and-association strategies that mainly rely on RGB information face clear limitations in occluded scenes(see Fig.~\ref{fig:intro}). Therefore, the core challenge of this task lies not only in cross-frame association and deduplication at the video level, but also in accurately distinguishing neighboring objects with highly ambiguous appearances within each frame.

To overcome the limitations of RGB features in  crowded and occluded scenes, introducing depth information to provide front-back relations and relative distance is an effective solution. For tightly arranged or mutually occluded objects of the same category, they still exhibit distinguishable depth differences and spatial hierarchy in 3D space. Such information can not only alleviate occlusion among adjacent objects within a single frame, but also provide geometric constraints that are less sensitive to appearance variation across frames. However, introducing depth does not automatically lead to consistent gains; the key lies in how to fuse it effectively. Direct channel concatenation of RGB and depth features, as shown in Fig. 1(b), ignores the distribution gap between the two modalities and therefore leads to many missed detections. Conventional early-fusion or late-fusion strategies also often suffer from insufficient feature interaction. Simple fusion methods usually fail to establish effective complementarity between the two modalities and cannot emphasize truly useful geometric information in critical regions. Therefore, for object-level video counting to truly benefit from depth, the key lies not in simply introducing depth, but in designing an effective fusion mechanism that can better handle occlusion, appearance similarity, and multi-scale geometric variation.

To fully exploit the potential of depth information for this challenging counting task, we propose a unified framework consisting of depth-guided detection and depth-enhanced deduplication. First, in the detection stage, we design a Depth-Guided Detector (DG-Det). Since RGB appearance and depth have different representation properties, DG-Det introduces multi-scale RGB-D cross-attention with Depth Affinity Bias at each level of the feature pyramid. By imposing geometric constraints on feature interaction through local depth relations, it explicitly models the complementarity between the two modalities and promotes feature aggregation in geometrically consistent regions, thereby suppressing background interference and cross-level mismatching. In addition, we design an occlusion-aware prediction head to explicitly estimate the occlusion probability of each target, providing reliable confidence cues for subsequent counting. Second, in the cross-frame association stage, we propose Depth-Guided Tracking and De-duplication. We incorporate depth constraints between objects into the trajectory matching cost, which improves the robustness of cross-frame matching. On top of this, Occlusion-Adaptive Temporal Voting adaptively adjusts the temporal filtering threshold using the predicted occlusion probability, enabling the recovery of heavily occluded target trajectories while maintaining high precision. Furthermore, since existing video counting datasets contain only RGB information, we construct and release the RGBD-VideoCount dataset. This dataset provides synchronized RGB-D video streams and contains 195 video clips across 6 categories, covering complex scenarios. Experimental results show that the proposed method achieves significant improvements over existing baselines on RGBD-VideoCount. Our contributions are as follows:

\begin{itemize}
    \item We introduce depth cues into video individual counting in crowded scenes, effectively alleviating the limitations of RGB-only methods in occluded scenarios.
    \item We propose a depth-guided RGB-D detection and de-duplication framework that better distinguishes occluded instances while reducing missed detections and duplicate counting.
    \item We release an RGBD-VideoCount dataset with synchronized depth information and multi-category coexisting scenes to facilitate research on video counting with depth.
\end{itemize}

\section{Related Work}
This section reviews the development of object counting from static images to videos, and further discusses the role of depth information in detection and tracking frameworks.

Existing image counting methods can be broadly divided into two categories: density map regression methods~\cite{arteta2014interactive,kong2006viewpoint,cho1999neural,liu2024weakly} and detection-based instance counting methods~\cite{barinova2010detection,desai2011discriminative,nguyen2022few,liu2024consistency,liu2021exploiting}. The former perform well in highly crowded and homogeneous scenes, but the category is usually implicitly encoded in the network parameters, which limits their generalization to unseen categories and prevents instance-level localization~\cite{gao2020cnn,song2021rethinking}. The latter provides explicit localization, but most of them rely on predefined categories and often degrade in crowded scenes~\cite{wu2023boosting,li2021approaches}. To overcome the limitation of closed-set categories, open-world counting has received increasing attention. Few-shot counting uses visual exemplars for matching, but it is sensitive to exemplar quality and appearance variation~\cite{djukic2023low,gong2022class,liu2022countr}. Zero-shot counting leverages the semantic transfer ability of vision-language models, but it is still easily affected by semantic ambiguity on fine-grained targets~\cite{amini2023open,dai2024referring,jiang2023clip}. 
Recent works unify multi-modal prompts for open-world counting, either through density estimation (Lin et al. \cite{lin2024fixed}) or detection-based frameworks (CountGD~\cite{amini2024countgd}). Related research in the video domain remains limited. As a representative method, CountVID~\cite{amini2026open} primarily enhances temporal counting uniqueness via promptable segmentation and pseudo-track filtering. However, in real-world occluded scenes, the high visual similarity and severe spatial overlap among objects make this approach highly susceptible to missed detections and track interruptions. Unlike previous video object counting methods, our approach innovatively incorporates depth to mitigate severe occlusion challenges faced by models that rely solely on 2D RGB features. By leveraging spatial hierarchy information, it achieves more accurate separation of visually overlapping instances.

In single-frame visual perception, depth has been introduced into detection frameworks through RGB-D encoding, feature interaction, or attention recalibration to enhance spatial modeling and occlusion awareness. For example, DGT~\cite{xu2025dgt} uses a Transformer to explicitly model depth-guided occlusion awareness for RGB-D occluded object detection. MonoDETR\cite{zhang2023monodetr} further reformulates the Transformer into a depth-aware structure, allowing object queries to interact with scene depth and thus reducing reliance on local appearance cues. Unlike existing depth-aware detectors dedicated to 3D pose regression or salient object extraction, our depth-guided detector not only uses depth features to separate overlapping targets within a single frame, but also explicitly predicts occlusion to provide crucial cues for cross-frame trajectory de-duplication,making it highly suitable for the VIC task in real-world occluded scenes

In video tasks, recent studies have started to use depth in the design of association costs, trajectory refinement, and occlusion reasoning for multi-object tracking, in order to improve temporal stability in complex scenes. For example, DepthMOT\cite{wu2024depthmot}uses depth cues and camera pose estimation to reduce ID switches caused by occlusion and camera motion. DepTR-MOT\cite{deng2025deptr} introduces instance-level depth supervision and depth consistency into a tracking-by-detection framework, thereby improving trajectory robustness under occlusion and close-range interactions.Although both MOT and our method use depth to reduce occlusion, MOT focuses on long-term ID tracking in real-world complex scenes, whereas video counting focuses on cross-frame de-duplication. Therefore, our approach, which effectively utilizes depth information during cross-frame association and dynamically evaluates trajectory validity via Occlusion-Adaptive Temporal Voting, is suited for this task.

\section{Method}
We consider a synchronized RGB-D video sequence $V = \{(I_t, Z_t)\}_{t=1}^T$, where $I_t$ and $Z_t$ denote the RGB image and the aligned depth map at frame $t$. Given an open-vocabulary text query $q$, a visual exemplar $VE$, and a video-level counting region (ROI), our goal is to output the individual number $C_{std}$ within the ROI. The proposed method consists of three components: a Depth-Guided  Detector, Depth-Guided Tracking and De-duplication, and training strategy.

\begin{figure*}[t]
    \centering
    \includegraphics[width=\textwidth]{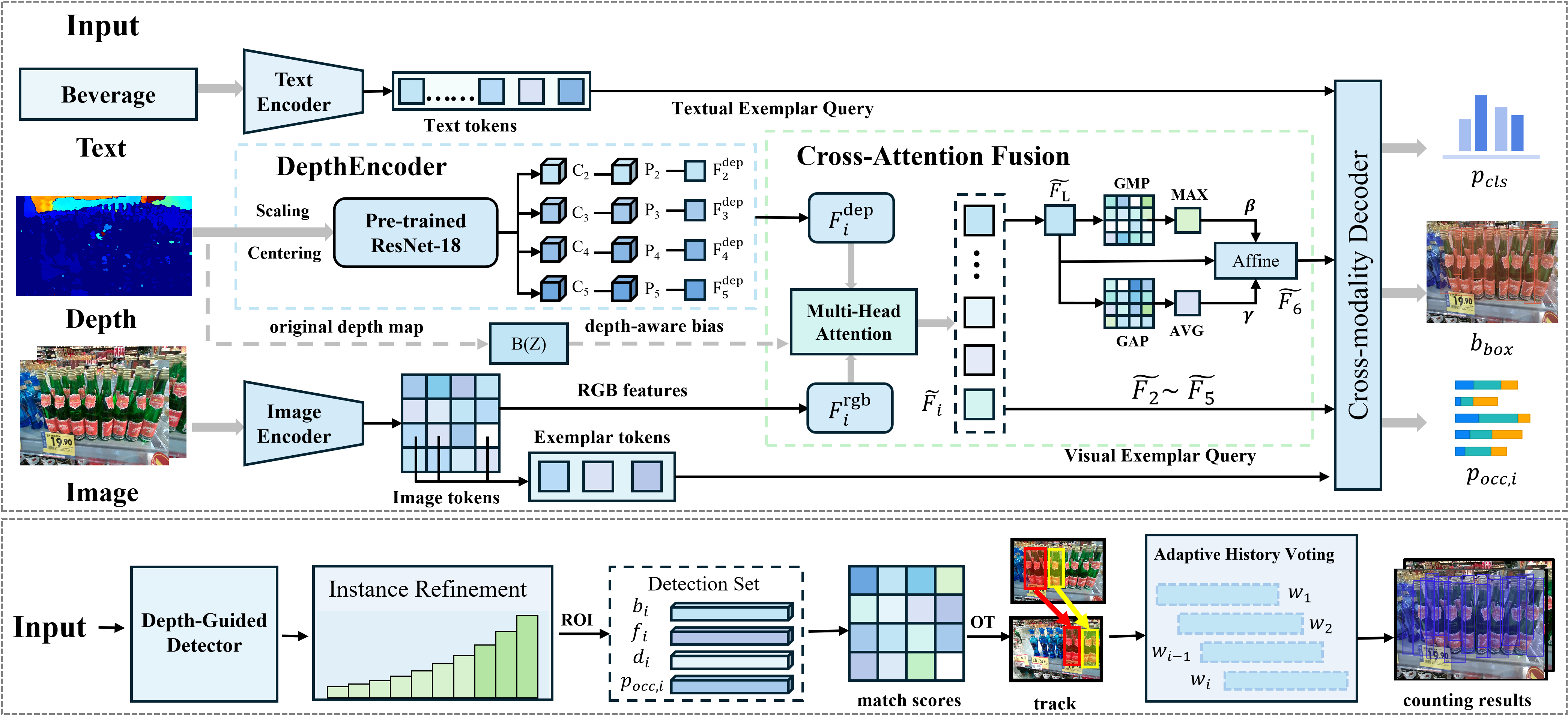} 
    \Description{Overall architecture of our framework. Top: the Depth-Guided Detector (DG-Det) takes a text and visual query and synchronized RGB–D inputs, fuses depth and RGB features via cross-modal attention, and predicts class scores, bounding boxes, and occlusion probabilities. Bottom: the Depth-Guided Tracking and De-duplication(DG-Track) module refines detections within the ROI, performs depth-aware online association to form stable trajectories, and applies adaptive history-window voting to remove duplicates and obtain consistent video counting results.}
    \caption{Overall architecture of our framework. Top: the Depth-Guided Detector (DG-Det) takes a text and visual query and synchronized RGB–D inputs, fuses depth and RGB features via cross-modal attention, and predicts class scores, bounding boxes, and occlusion probabilities. Bottom: the Depth-Guided Tracking and De-duplication(DG-Track) module refines detections within the ROI, performs depth-aware online association to form stable trajectories, and applies adaptive history-window voting to remove duplicates and obtain consistent video counting results.}
    \label{fig:overview}
\end{figure*}

\subsection{Depth-Guided Detector}
We begin by encoding the input query $q$ with a BERT encoder, and retain the tokens corresponding to the category name to guide cross-modal decoding. On the visual side, we adopt the backbone and FPN structure of GroundingDINO\cite{liu2024grounding} to extract multi-scale RGB features $\{F_{i}^{\text{rgb}}\}_{i=2}^5$ from $I_t$ as visual inputs to the Transformer. To incorporate depth cues aligned with the RGB feature space, we design a depth encoder $E_{\text{dep}}$. This encoder produces a depth pyramid that matches the multi-scale RGB features of GroundingDINO. Its architecture is based on ResNet-18 + FPN \cite{he2016deep,lin2017feature}, and the first convolutional layer is modified to accept single-channel input, using the channel-wise mean of the RGB pretrained weights for initialization. Through top-down feature fusion and smoothing convolutions, we obtain a four-level depth feature pyramid $\{F_{i}^{\text{dep}}\}_{i=2}^5$.

Based on the aligned RGB and depth feature pyramids, we design a Cross-Attention Fusion module to inject depth cues into the visual representation. Specifically, at each scale $i$, we project the RGB feature $F_{i,t}^{\text{rgb}}$ into queries $Q$, and the depth feature $F_{i,t}^{\text{dep}}$ into keys $K$ and values $V$. To suppress spurious associations between regions that are distant in 3D space but appear nearby in the image plane, we introduce a Depth Affinity Bias $B(Z_t)$ into the attention computation. The resulting attention weights are defined as
\begin{equation}
\alpha_{uv}^{(i,t)}=
\frac{
\exp\left(\frac{q_u k_v^\top}{\sqrt d}+B_{uv}(Z_t)\right)
}{
\sum_{v'}
\exp\left(\frac{q_u k_{v'}^\top}{\sqrt d}+B_{uv'}(Z_t)\right)
},
\label{eq:attn_depth_bias}
\end{equation}
where  Depth Affinity Bias is given by 
\begin{equation}
B_{uv}(Z_t) = -\beta_{\text{bias}} \cdot \bigl| \phi(Z_t)_u - \phi(Z_t)_v \bigr|,
\label{eq:depth_affinity}
\end{equation}
where $\phi(\cdot)$ is a depth normalization function (including log-inverse transformation and quantile normalization), and $\beta_{\text{bias}}$ is a coefficient controlling the strength of depth suppression. Based on these weights, the depth-guided fusion feature at scale $i$ is computed as 
\begin{equation}
\hat{F}_{i,t}(u)=\sum_v \alpha_{uv}^{(i,t)}v_v,
\end{equation}
In this way, feature interaction between regions with different depth values is effectively suppressed. The final multi-scale fused features $\{\tilde{F}_{i,t}\}_{i=2}^5$ thus encode both RGB and depth information.

To address the mismatch between the Transformer's requirement for four input scales and the backbone's output of only three, previous methods often use simple downsampling, which lacks global depth information. To overcome this limitation, we propose a global feature generation strategy based on Feature-wise Linear Modulation (FiLM)\cite{perez2018film}. SE blocks mainly perform multiplicative reweighting, while conditional normalization depends on normalization statistics and may disturb the pretrained feature distribution. In contrast, FiLM enables conditional affine modulation through both scaling and shifting, making it more suitable for injecting global depth priors to generate the additional coarse-scale feature.

Specifically, we apply global average pooling (GAP) and global max pooling (GMP) on the depth map $Z_t$ to extract scene-level depth statistics, and use $1\times1$ convolutions to generate channel-wise scaling and shifting parameters $\gamma$ and $\beta$:
\begin{equation}
\gamma = \sigma\bigl(W_\gamma \cdot \text{GAP}(Z_{L,t})\bigr), \quad
\beta = W_\beta \cdot \text{GMP}(Z_{L,t}).
\label{eq:film_params}
\end{equation}
We then perform a channel-wise affine transformation on the top-level fused feature $\tilde{F}_{L,t}$, followed by downsampling to generate the fourth-level feature $\tilde{F}_{6}$. Finally, we feed $\{\tilde{F}_2,\tilde{F}_3,\tilde{F}_4,\tilde{F}_5,\tilde{F}_6\}$ into the subsequent multi-layer Transformer.

Building upon the fused multi-scale features, we further introduce an occlusion-aware design to explicitly model dense-scene occlusion. Specifically, we generate a continuous occlusion score $o_i \in [0,1]$ for each ground-truth object during data preprocessing, rather than using a simple binary label. Let $B_i$ denote the bounding box of the $i$-th object. If a neighboring object spatially overlaps with $B_i$ and is closer to the camera in depth, it is considered to contribute to the occlusion of object $i$. We jointly compute the occlusion score $o_i$ according to both the spatial coverage of neighboring objects over object $i$ and their relative depth relationships, where a larger $o_i$ indicates a higher degree of foreground occlusion. The detailed construction of $o_i$ is provided in the supplementary material.

Through the above process, we build a detector that jointly models semantic, depth, and occlusion information, providing reliable multi-modal foundation for subsequent video counting tracks.

\subsection{Depth-Guided Tracking and De-duplication}
Given a video sequence $V$ and an open-vocabulary query set $q$, we first run our DG-Det on each frame to obtain the raw detection set $\tilde{\mathcal{D}}_t$. These are then sequentially processed by detection refinement, online tracking, and temporal voting to produce the final count:
\begin{equation}
\tilde{\mathcal{D}}_t
\xrightarrow{\ \mathcal{F}_{\text{det}}\ }
\mathcal{D}_t
\xrightarrow{\ \mathcal{F}_{\text{trk}}\ }
\mathcal{T}
\xrightarrow{\ \mathcal{F}_{\text{cnt}}\ }
C_{\text{std}},
\label{eq:pipeline}
\end{equation}
where $\mathcal{D}_t$ is the refined detection set at frame $t$, $\mathcal{T}=\{T_k\}$ is the set of video-level object trajectories, and $C_{\text{std}}$ is the final count. More details are provided in
the Supplementary Material.

Starting from the raw detections produced by DG-Det, we first perform per-frame detection refinement to obtain reliable instance candidates for subsequent association. For each frame and each query category, we first compute the phrase-level confidence score $S$ from the logits and obtain the corresponding occlusion probability $P_{\text{occ}}$ from the decoder. To avoid losing potential targets at the early filtering stage, we adopt a two-stage filtering strategy: 1) Candidate boxes satisfying $S > \tau_{c}$ are retained.
2) Post-NMS, candidates are spatially filtered to retain those exceeding an overlap threshold with the video-level ROI.
Finally, for each remaining candidate, we extract its bounding box $b_i$, decoder feature vector $f_i$, sampled depth $d_i$, and occlusion probability $p_{\text{occ},i}$, which together form the detection set $\mathcal{D}_t$ of the current frame.

To remove duplicate counts across frames, we perform online multi-object tracking based on ObjectTracker. Unlike conventional multi-object tracking methods that rely solely on 2D positions and appearance cues, we incorporate depth into trajectory association, yielding depth-guided association and de-duplication. For a trajectory $tr_i \in T_{t-1}$ and a detection $det_j \in D_t$, let $\Delta \mathbf{p}_{ij}$ denote the center displacement, $\Delta z_{ij}$ denote the center-depth difference, and $\mathbf{u}_i,\mathbf{u}_j$
denote the normalized appearance features. We model the matching event $M_{ij}=1$ using the following modality-specific likelihoods, whose motivation and detailed derivation are provided in the Supplementary Material:
\begin{equation}
\begin{aligned}
p(\Delta \mathbf{p}_{ij}\mid M_{ij}=1)
&\propto
\exp\!\left(-\frac{\|\Delta \mathbf{p}_{ij}\|_2}{\sigma_p}\right), \\
p(\Delta z_{ij}\mid M_{ij}=1)
&\propto
\exp\!\left(-\frac{|\Delta z_{ij}|}{\sigma_d}\right), \\
p(\mathbf{u}_j \mid \mathbf{u}_i, M_{ij}=1)
&\propto
\exp\!\left(\kappa_f\,\mathbf{u}_i^\top \mathbf{u}_j\right).
\end{aligned}
\end{equation}
where $\sigma_p$ and $\sigma_d$ denote the effective noise scales of the position and depth modalities, respectively, and $\kappa_f$ denotes the concentration of appearance consistency.
Assuming conditional independence across modalities given the matching event, and introducing an IoU-based geometric prior $p_{\mathrm{geo}}(M_{ij}=1)$, the matching posterior can be written as
\begin{equation}
\begin{aligned}
p(M_{ij}=1 \mid tr_i, det_j)
\propto\;&
p(\Delta \mathbf{p}_{ij}\mid M_{ij}=1)\,
p(\Delta z_{ij}\mid M_{ij}=1) \\
&\cdot\,
p(\mathbf{u}_j\mid \mathbf{u}_i, M_{ij}=1)\,
p_{\mathrm{geo}}(M_{ij}=1),
\end{aligned}
\end{equation}
We then define the association cost by minimizing the negative log-posterior:
\begin{equation}
\begin{split}
C_{ij}
&= -\log p(M_{ij}=1 \mid tr_i, det_j) \\
&= w_p\, d_{\mathrm{pos}}(tr_i,det_j)
 +  w_f\, d_{\mathrm{feat}}(tr_i,det_j) 
+ w_d\, d_{\mathrm{depth}}(tr_i,det_j)\\
&\quad + \lambda_{\mathrm{iou}}\,
I\!\bigl(\mathrm{IoU}(b_i,b_j)<\tau_{\mathrm{iou}}\bigr)
 + \mathrm{const},
\end{split}
\raisetag{11pt}
\end{equation}
where $d_{\mathrm{pos}}$ is the Euclidean distance between box centers,
$d_{\mathrm{feat}}$ is the cosine distance between appearance features, and
$d_{\mathrm{depth}}$ is the center-depth difference. The weights are related to
modality reliability as
\begin{equation}
w_p \propto \sigma_p^{-1},\qquad
w_d \propto \sigma_d^{-1},\qquad
w_f \propto \kappa_f.
\end{equation}

We normalize the cost matrix, use the Hungarian algorithm to obtain the minimum-cost assignment, and accept only matched pairs whose cost is below a predefined threshold. Matched tracks are updated with the new bounding box, depth, and features; unmatched tracks are deactivated if they are not observed for a consecutive number of frames; unmatched detections are initialized as new tracks. Finally, we obtain a set of stable target trajectories
$\{\text{track}_k\}$.

To further suppress spurious trajectories caused by transient false positives and short-term occlusions, we perform temporal-consistency-based validity verification for each trajectory $T_k$. Specifically, we first compute the trajectory lifespan $L_k$, the confidence statistics $(S_{\max}, \bar{S})$, and the mean occlusion rate $\bar{O}_k$. After removing noisy trajectories with excessively short lifespans ($L_k < H_{\mathrm{base}}$) or low confidence, we adaptively determine the minimum hit count $N_{\min}$ and the frame-density threshold $\rho$ according to $\bar{O}_k$.Next, for each candidate trajectory, we slide a temporal window of length $W$ over its time span and count the number of frames $h$ in which the object is successfully detected within the window. Let $n_k=\min(W,L_k)$ denote the effective window length for the trajectory $T_k$. To jointly evaluate the absolute number of hits and the local hit density, we define a trajectory support score as
\begin{equation}
s_k=\min\left(\frac{h}{N_{\min}},\frac{h/n_k}{\rho}\right),
\end{equation}
Here, the first term measures whether the trajectory has accumulated a sufficient number of detections, while the second term evaluates whether its detections are temporally dense enough within the local window. A larger $s_k$ therefore indicates stronger temporal support for the trajectory. We regard a trajectory as valid if $s_k \ge 1$. The total number of valid trajectories then constitutes the final standard counting result $C_{\text{std}}$.

\subsection{Training}
To train the proposed model, we introduce the training strategy. In addition to the original classification and GIoU losses in CountGD\_Box, we further introduce a bounding-box regression loss and an occlusion prediction loss to enhance localization accuracy in overlapping regions and explicitly supervise the newly added occlusion branch.
Specifically, to further improve localization accuracy under dense overlap, we employ an L1 bounding-box regression loss:
\begin{equation}
\mathcal{L}_{\text{bbox}} = \frac{1}{N_{\text{pos}}} \sum_{i \in \text{Pos}} \bigl\lVert b_i - \hat{b}_i \bigr\rVert_1,
\label{eq:bbox_loss}
\end{equation}
where $N_{\text{pos}}$ is the number of matched boxes, and $b_i$ and $\hat{b}_i$ denote the predicted and ground-truth boxes.
To supervise the newly added occlusion prediction branch, we define the occlusion loss as
\begin{equation}
\mathcal{L}_{\text{occ}} = \frac{1}{N_{\text{pos}}} \sum_{i \in \text{Pos}} \text{BCE}(p_{\text{occ}, i}, o_i),
\label{eq:occ_loss}
\end{equation}
During training, the total loss $\mathcal{L}$ is given by
\begin{equation}
\mathcal{L} = \lambda_{1}\mathcal{L}_{\text{cls}} + \lambda_{2}\mathcal{L}_{\text{giou}} + \lambda_{3}\mathcal{L}_{\text{bbox}} + \lambda_{4}\mathcal{L}_{\text{occ}},
\label{eq:total_loss}
\end{equation}
where $\mathcal{L}_{\text{cls}}$ and $\mathcal{L}_{\text{giou}}$ are the classification loss and GIoU loss in CountGD\_Box, and $\lambda_{1}$, $\lambda_{2}$, $\lambda_{3}$, and $\lambda_{4}$ are hyper-parameters.
For optimization, we initialize the model from CountGD\_Box and freeze the pretrained backbone, while optimizing the newly introduced modules, including the depth encoder, the cross-attention fusion module, and the detection head. Training is conducted with AdamW using a batch size of 1, together with standard data augmentation such as scaling, cropping, and flipping.More details are provided in
the Supplementary Material.
\begin{table}[t]
\centering
\caption{RGBD-VideoCount Dataset information.}
\label{tab:dataset_overview}
\setlength{\tabcolsep}{4pt}
\renewcommand{\arraystretch}{1.08}
\small
\begin{tabular*}{\columnwidth}{@{\extracolsep{\fill}}lccccc@{}}
\toprule
Dataset & Cls. & Mod. & Vid. & MC & Ann. \\
\midrule
CARPK\cite{hsieh2017drone}      & 1   & RGB   & N & -- & B \\
COCO\cite{lin2014microsoft}     & 80  & RGB   & N & -- & B, IM \\
FSC147\cite{ranjan2021learning} & 147 & RGB   & N & -- & P, EB \\
VideoCount\cite{amini2026open}  & 141 & RGB   & Y & -- & B, EB \\
\textbf{RGBD-VideoCount}        & \textbf{6} & \textbf{RGB-D} & \textbf{Y} & \textbf{Y} & \textbf{B, EB, ROI} \\
\midrule
\multicolumn{6}{@{}l}{\textit{RGBD-VideoCount category information}} \\
\midrule
Category & \#Vid. & Min & Max & \multicolumn{2}{c}{Avg.} \\
\midrule
beverage  & 94 & 1  & 208 & \multicolumn{2}{c}{48.33} \\
bag       & 8  & 3  & 23  & \multicolumn{2}{c}{13.50} \\
book      & 42 & 28 & 209 & \multicolumn{2}{c}{114.20} \\
plant     & 12 & 5  & 93  & \multicolumn{2}{c}{25.92} \\
box       & 55 & 1  & 143 & \multicolumn{2}{c}{20.42} \\
cabinet   & 42 & 12 & 132 & \multicolumn{2}{c}{69.86} \\
\midrule
Overall & 195+ & 1 & 209 & \multicolumn{2}{c}{70.51} \\
\bottomrule
\end{tabular*}
\raggedright
\footnotesize
B: boxes; P: points; EB: exemplar boxes; IM: instance masks; ROI: region of interest; 
MC: multiple object categories may coexist in one video. 
\end{table}
\section{RGBD-VideoCount Dataset}
Most existing video counting datasets only provide RGB modality, which makes it difficult to explicitly model occlusion relationships(see Table~\ref{tab:dataset_overview}). To address this limitation, we release \textbf{RGBD-VideoCount}, a synchronized RGB-D dataset for video individual counting in crowded scenes. Unlike previous datasets that only provide RGB videos and usually focus on a single target category, RGBD-VideoCount allows multiple categories to coexist in the same video, and also provides depth information, instance-level bounding box annotations, exemplar boxes, and video-level ROIs. This makes it a unified benchmark for depth-assisted video detection, cross-frame association, and de-duplication. Such a design is closer to real shelf and inventory scenarios, and it also provides a more effective testbed for evaluating the robustness of counting methods in the presence of distracting categories. 

RGBD-VideoCount contains \textbf{2,032} finely annotated frames and \textbf{77,638} instance boxes. The number of finely annotated frames exceeds that of CARPK, while the number of instance boxes is of a similar scale. The data distribution is designed to be close to real-world scenes. Specifically, the densely arranged Beverage category serves as a challenging setting for heavy occlusion, while categories such as Book, which have relatively weaker occlusion but still dense layouts, are used to evaluate the model’s generalization ability in high-density localization scenarios. The data are collected using an \textbf{Intel RealSense D455}, with the RGB and depth streams strictly synchronized through SDK-level alignment. More details are provided in
the Supplementary Material.

\begin{table*}[t]
\centering
\scriptsize
\setlength{\tabcolsep}{3.5pt}
\caption{Results on RGBD-VideoCount. Methods marked with $^\dagger$ utilize estimated depth.}
\label{tab:counting-results}
\resizebox{\textwidth}{!}{
\begin{tabular}{c c c c c c c c c c c}
\toprule
\multirow{3}{*}{Method} & \multirow{3}{*}{Detector} & \multirow{3}{*}{Tracker} & 
\multicolumn{4}{c}{RGBD-VideoCount Val} & \multicolumn{4}{c}{RGBD-VideoCount Test} \\
\cmidrule(lr){4-7} \cmidrule(lr){8-11}
& & & \multicolumn{2}{c}{Counting} & \multicolumn{2}{c}{Detection} & \multicolumn{2}{c}{Counting} & \multicolumn{2}{c}{Detection} \\
\cmidrule(lr){4-5} \cmidrule(lr){6-7} \cmidrule(lr){8-9} \cmidrule(lr){10-11}
& & & MAE$\downarrow$ & RMSE$\downarrow$ & AP$\uparrow$ & AP50$\uparrow$ & MAE$\downarrow$ & RMSE$\downarrow$ & AP$\uparrow$ & AP50$\uparrow$ \\
\midrule

\multicolumn{11}{c}{\textit{Text-only prompt}} \\
\midrule
CountVID & COUNTGD-BOX & SAM~2.1
& 24.77 & 28.76 & 11.59 & 25.76
& 23.69 & 30.78 & 9.38 & 23.21 \\
-- & COUNTGD-BOX & ByteTrack
& 28.78 & 43.5 & 4.44 & 13.54
& 25.74 & 37.28 & 4.87 & 15.19 \\
-- & COUNTGD-BOX & OC-SORT
& 34.74 & 55.78 & 6.81 & 18.17
& 22.29 & 36.34 & 7.40 & 18.85 \\
-- & COUNTGD-BOX & SparseTrack$^\dagger$
& 26.74 & 42.11 & 5.89 & 16.89
& 24.86 & 36.51 & 7.06 & 20.31 \\
Ours$^\dagger$ & DG-Det$^\dagger$ & DG-Track$^\dagger$
& 16.95 & 28.39 & \textbf{21.95} & \textbf{45.16}
& 13.75 & 20.81 & 16.97 & 44.79 \\
\textbf{Ours} & \textbf{DG-Det} & \textbf{DG-Track}
& \textbf{14.55} & \textbf{25.32} & 20.16 & 43.26
& \textbf{12.68} & \textbf{15.87} & \textbf{17.52} & \textbf{45.68} \\
\midrule

\multicolumn{11}{c}{\textit{Text + exemplar prompt}} \\
\midrule
CountVID & COUNTGD-BOX & SAM~2.1
& 19.68 & 26.87 & 18.73 & 47.65
& 20.98 & 27.73 & 17.08 & 48.42 \\
-- & COUNTGD-BOX & ByteTrack
& 25.51 & 37.83 & 11.32 & 32.25
& 23.17 & 31.95 & 11.55 & 34.49 \\
-- & COUNTGD-BOX & OC-SORT
& 38.41 & 55.57 & 13.89 & 30.29
& 25.26 & 38.89 & 13.35 & 31.68 \\
-- & COUNTGD-BOX & SparseTrack$^\dagger$
& 25.22 & 37.47 & 16.26 & 39.03
& 23.02 & 32.17 & 15.75 & 42.84 \\
Ours$^\dagger$ & DG-Det$^\dagger$ & DG-Track$^\dagger$
& 14.13 & 28.21 & 22.50 & 48.99
& 13.63 & 33.51 & \textbf{22.92} & \textbf{53.00} \\
\textbf{Ours} & \textbf{DG-Det} & \textbf{DG-Track}
& \textbf{13.32} & \textbf{19.68} & \textbf{23.96} & \textbf{50.12}
& \textbf{7.97} & \textbf{11.44} & 20.78 & 50.39 \\
\midrule

\multicolumn{11}{c}{\textit{RGB-D tracking Model}} \\
\midrule
DepthMOT$^\dagger$ & -- & --
& 26.38 & 43.53 & 2.80 & 7.87
& 23.57 & 32.65 & 4.08 & 9.30 \\
DepTR-MOT$^\dagger$ & -- & --
& 18.67 & 41.27 & 3.05 & 11.47
& 14.33 & 25.75 & 3.40 & 13.28 \\
\textbf{Ours} & \textbf{DG-Det} & \textbf{DG-Track}
& \textbf{13.32} & \textbf{19.68} & \textbf{23.96} & \textbf{50.12}
& \textbf{7.97} & \textbf{11.44} & \textbf{20.78} & \textbf{50.39} \\
\bottomrule
\end{tabular}
}
\end{table*}

\section{Experiments}
We conduct comprehensive experiments to evaluate the effectiveness of our DG-Det and DG-Track. First, we introduce the evaluation metrics. Then, we compare our method with the baseline and several state-of-the-art approaches. Finally, we perform ablation studies to analyze the contribution of each key component.

\subsection{Metrics}
We adopt both counting and localization metrics to evaluate model performance. For each video $i$, we compute the discrepancy between the predicted count $\hat{y}_i$ and the ground-truth count $y_i$. We use the Mean Absolute Error (MAE) and the Root Mean Squared Error (RMSE) to measure counting accuracy. In addition, we evaluate localization performance using Average Precision (AP) and $\mathrm{AP}_{50}$, where $\mathrm{AP}_{50}$ denotes the AP computed at an IoU threshold of 0.5.

\subsection{Comparison with Existing Methods}
We first compare our method with CountVID on RGBD-VideoCount, and construct three representative RGB tracking models based on CountGD-Box, namely ByteTrack, OC-SORT, and SparseTrack. ByteTrack and OC-SORT emphasize low-confidence detection recovery and motion modeling, while SparseTrack performs cascaded matching using pseudo-depth estimated from single-frame RGB images. Together, these models cover representative temporal association strategies commonly used in crowded video counting.

\begin{figure}[t]
    \centering
    \includegraphics[width=\linewidth]{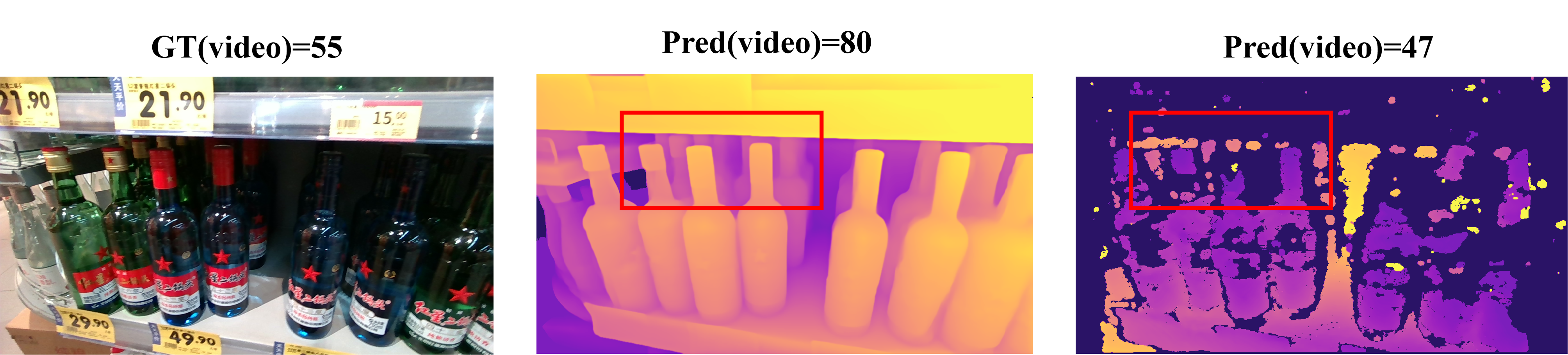}
    \Description{Qualitative comparison between predicted pseudo-depth (middle) and ground-truth depth (right)}
    \caption{Qualitative comparison between predicted pseudo-depth (middle) and ground-truth depth (right).}
    \label{fig:depth}
\end{figure}

\begin{figure*}[t]
    \centering
    \includegraphics[width=\textwidth]{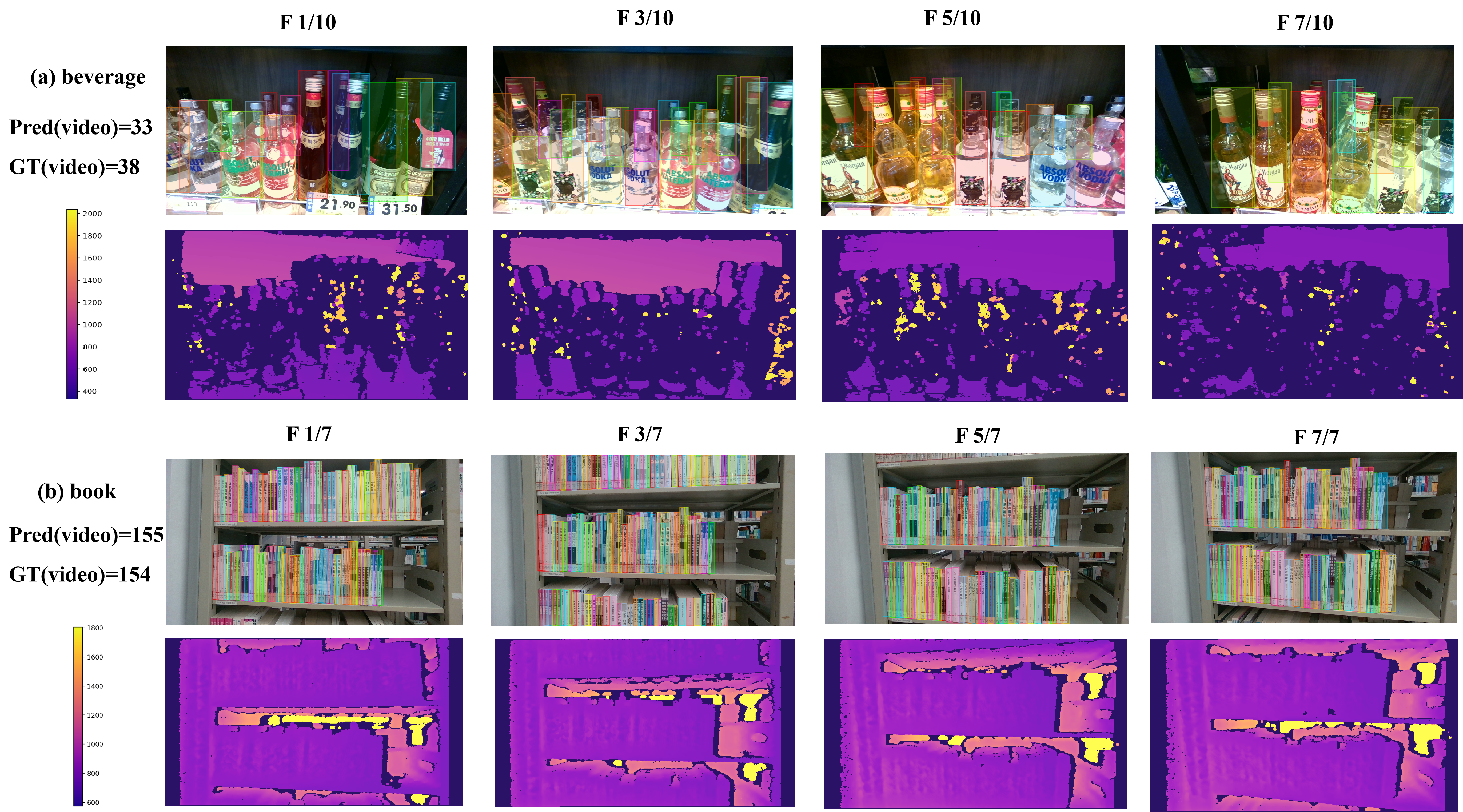}
    \Description{Example predictions of our method in two typical challenging scenarios. For each video, several sampled frames and their corresponding depth maps are shown, together with the video-level prediction Pred(video) and the ground-truth count GT(video). (a) A shelf video with severe occlusions (testing robustness to occlusion); (b) a bookshelf video without obvious occlusion but with highly crowded objects (testing localization accuracy in dense scenes). More scene details are provided in the Supplementary Material.}
    \caption{Example predictions of our method in two typical challenging scenarios. For each video, several sampled frames and their corresponding depth maps are shown, together with the video-level prediction Pred(video) and the ground-truth count GT(video). (a) A shelf video with severe occlusions (testing robustness to occlusion); (b) a bookshelf video without obvious occlusion but with highly crowded objects (testing localization accuracy in dense scenes). More scene details are provided in the Supplementary Material.}
    \label{fig:fig7}
\end{figure*}
Under the challenging text-only setting, pure RGB baselines generally suffer from severe missed detections and false positives due to the lack of visual exemplars, leading to poor localization accuracy and large counting errors. In contrast, our method shows a clear advantage (see Table~\ref{tab:counting-results}). This shows that text-only 2D features struggle in crowded scenes. By fusing multimodal features with spatial depth, our DG-Det achieves reliable localization for accurate counting. With the introduction of visual exemplars, the overall performance of all methods improves, and our approach still achieves the best results in both counting and detection metrics. Notably, the SAM 2.1 baseline achieves an excellent AP50 but a high MAE, showing that strong single-frame detection cannot ensure accurate counting under severe occlusion. Without explicit occlusion modeling, frequent ID switches and fragmented trajectories lead to severe over-counting. Our method avoids this by unifying precise localization with occlusion-guided temporal de-duplication.

To further verify that the gain of our method does not simply come from introducing an additional depth modality, we also compare with two RGB-D tracking methods (see Table~\ref{tab:counting-results}), namely DepthMOT\cite{wu2024depthmot} and DepTR-MOT\cite{deng2025deptr}. The results show that DepTR-MOT effectively utilizes depth information in both the detection and tracking stages to outperform pure RGB baselines. In contrast, DepthMOT exploits depth solely during the tracking stage, resulting in overall performance comparable to pure RGB trackers.  Furthermore, both tracking methods yield lower AP scores, suggesting that directly applying traditional tracking paradigms is challenging to maintain high localization accuracy under severe occlusions. In contrast, our method achieves the best performance under the same input setting. This result suggests that the key to the improvement lies not in naively adding depth, but in effectively modeling depth at the feature representation level and jointly exploiting occlusion cues during both detection and de-duplication.

To investigate the impact of depth quality, we evaluated our model using pseudo-depth ( Table~\ref{tab:counting-results}, Ours$\dagger$). As shown in the red box (Fig~\ref{fig:depth} ), the predicted depth maps suffer from "edge smoothing," erasing spatial boundaries between overlapping objects. In contrast, although the ground-truth depth is sparse and noisy, it keeps sharp depth changes. This provides more accurate geometric cues for distinguishing crowded objects, leading to better performance.

We further evaluate our method on FSCD-147\cite{nguyen2022few} using the pseudo-depth estimated by Depth Anything V2\cite{yang2024depth}. Details and qualitative results are provided in the supplementary material.

\subsection{Ablation Studies}

\begin{table}[t]
    \centering
    \caption{Ablation study on depth coverage.}
    \label{tab:depth_coverage}
    \scriptsize
    \setlength{\tabcolsep}{2.4pt}
    \renewcommand{\arraystretch}{0.92}
    \resizebox{\linewidth}{!}{
        \begin{tabular}{lcccc}
            \toprule
            \multirow{2}{*}{Method} &
            \multicolumn{2}{c}{Validation} &
            \multicolumn{2}{c}{Test} \\
            \cmidrule(lr){2-3} \cmidrule(lr){4-5}
            & MAE$\downarrow$ & RMSE$\downarrow$
            & MAE$\downarrow$ & RMSE$\downarrow$ \\
            \midrule
            RGB-only (0\%)
            & 19.88 & 28.46 & 15.47 & 25.26 \\
            Partial depth (25\%)
            & $16.15 \pm 0.29$ & $30.25 \pm 1.57$
            & $14.74 \pm 0.17$ & $35.85 \pm 0.82$ \\
            Partial depth (50\%)
            & $16.52 \pm 0.40$ & $28.87 \pm 0.61$
            & $12.60 \pm 0.36$ & $19.66 \pm 0.93$ \\
            Partial depth (75\%)
            & $14.18 \pm 0.27$ & $29.13 \pm 1.10$
            & $12.91 \pm 0.21$ & $36.44 \pm 0.82$ \\
            \textbf{Ours (100\%)}
            & \textbf{13.33} & \textbf{19.68}
            & \textbf{7.98} & \textbf{11.45} \\
            \bottomrule
        \end{tabular}
    }
    \vspace{-5pt}
\end{table}

\textbf{Ablation Study on depth coverage.}
Table~\ref{tab:depth_coverage} compares different depth coverage settings. For the 25\%, 50\%, and 75\% partial-depth settings, we report the mean and standard deviation over multiple random seeds. Compared with the RGB-only baseline, depth cues consistently reduce MAE, demonstrating their value for providing spatial constraints in crowded scenes. Partial depth improves MAE but can lead to RMSE fluctuations due to cross-modal inconsistency, while full real depth achieves the best and most stable results.

\begin{table}[t]
    \centering
    \caption{Fine-grained ablation of the proposed components. }
    \label{tab:module_ablation}
    \small
    \setlength{\tabcolsep}{4.2pt}
    \renewcommand{\arraystretch}{1.05}
    \begin{tabular}{lcccc}
        \toprule
        \multirow{2}{*}{Configuration} &
        \multicolumn{2}{c}{Validation} &
        \multicolumn{2}{c}{Test} \\
        \cmidrule(lr){2-3} \cmidrule(lr){4-5}
        & MAE$\downarrow$ & RMSE$\downarrow$
        & MAE$\downarrow$ & RMSE$\downarrow$ \\
        \midrule
        RGB-only
        & 18.72 & 32.40 & 18.07 & 26.80 \\
        + Cross-Attention Fusion
        & 17.12 & 28.77 & 14.93 & 20.89 \\
        + Depth Affinity Bias
        & 14.26 & 26.08 & 12.20 & 15.99 \\
        + Occlusion Head
        & \textbf{13.33} & \textbf{19.68}
        & \textbf{7.98} & \textbf{11.45} \\
        \bottomrule
    \end{tabular}
\end{table}

\textbf{Fine-grained Module Ablation.}
 To further analyze the contribution of each component, we start from the RGB-only baseline and progressively add Cross-Attention Fusion, Depth Affinity Bias, and the Occlusion Head, as shown in Table~\ref{tab:module_ablation}. Adding Cross-Attention Fusion reduces the validation and test MAE/RMSE from 18.72/32.40 and 18.07/26.80 to 17.12/28.77 and 14.93/20.89, respectively, demonstrating that depth feature fusion improves target localization in crowded scenes. With Depth Affinity Bias, the results further improve to 14.26/26.08 on validation and 12.20/15.99 on test, indicating more effective use of local depth relations. Finally, adding the Occlusion Head achieves the best performance, with MAE/RMSE of 13.33/19.68 on validation and 7.98/11.45 on test. These progressive gains show that the proposed components jointly improve target reliability estimation and cross-frame de-duplication .

\subsection{Robustness and Sensitivity Analysis}

We first analyze the sensitivity of the depth affinity bias coefficient $\beta$. As shown in the table~\ref{tab:ablation_robustness}, when $\beta$ varies within the range of 0.5--3.0, the model performance remains stable, with only minor fluctuations in MAE and RMSE. This indicates that the proposed \textit{Depth Affinity Bias} maintains good robustness across a relatively wide range of parameter values. Overall, the introduced bias enables the model to exploit depth hierarchy information more consistently, leading to improved detection and counting performance in complex scenes.

We further evaluate the effectiveness of \textit{Occlusion-Adaptive Temporal Voting}. Specifically, we compare the adaptive voting strategy with a fixed-threshold temporal filtering scheme. As shown in Table~\ref{tab:ablation_robustness}, introducing the adaptive voting mechanism reduces MAE and RMSE by 1.91 and 2.75, respectively. These results show that \textit{Occlusion-Adaptive Temporal Voting} achieves a better balance between suppressing spurious trajectories and recovering heavily occluded targets. Compared with a fixed-threshold scheme, our method dynamically adjusts the trajectory retention criterion according to the predicted occlusion probability, leading to more stable and accurate video-level counting in occlusion scenarios.

Considering that depth sensors in real-world applications may suffer from noise, blur, and missing values, we further corrupt the test-set depth maps with several perturbations, including \textit{Gaussian noise}, \textit{Salt \& Pepper noise}, \textit{Edge Blur}, and \textit{Random Holes}. The results show that our method remains highly robust under \textit{Edge Blur} and \textit{Random Holes}, achieving test MAEs of 8.50 and 8.23, respectively, which are close to the 7.97 obtained with clean depth input. In contrast, both \textit{Gaussian noise} and \textit{Salt \& Pepper noise} lead to noticeable performance degradation, with the MAE increasing to 11.16 and 10.66, respectively. We attribute this to the fact that both types of noise directly corrupt local depth structures and fine-grained depth ordering, thereby reducing the effectiveness of the bias in cross-modal fusion. Overall, these experiments suggest that our method is more robust to structured depth degradation such as blur and partial missing regions, while random noise remains more challenging for reliable counting.
\begin{table}[t]
  \centering
  \caption{Ablation Study and Robustness Analysis.}
  \resizebox{0.48\textwidth}{!}{
  \begin{tabular}{l c c c c c c}
    \toprule
    \textbf{Setting / Condition} & \textbf{Param.} & \textbf{SNR (dB)} & \textbf{MAE} & \textbf{RMSE} & \textbf{$\Delta$MAE} & \textbf{$\Delta$RMSE} \\
    \midrule
    \multicolumn{7}{l}{\textit{Experimental Results on Depth Noise Robustness}} \\
    \midrule
    ours (clean) & -- & -- & \textbf{7.97} & \textbf{11.44} & -- & -- \\
    Gaussian  & $\sigma=0.05$ & 10.2 & 11.16 & 21.13 & +3.18 & +9.68 \\
    Salt \& Pepper & ratio=5\% & -0.4 & 10.66 & 13.31 & +2.68 & +1.86 \\
    Edge Blur& k=$15\times15$ & 15.2 & 8.50 & 12.46 & +0.52 & +1.00 \\
    Random Holes  & area$\approx 5\%$ & 18.5 & 8.23 & 12.11 & +0.25 & +0.66 \\
    \midrule
    \multicolumn{7}{l}{\textit{Sensitivity Analysis of Depth Bias}} \\
    \midrule
    $\beta=0.5$ & -- & -- & 8.02 & 11.69 & -0.05 & +0.24 \\
    \textbf{$\beta=1.0^\dagger$} & \textbf{--} & \textbf{--} & \textbf{7.97} & \textbf{11.44} & \textbf{--} & \textbf{--} \\
    $\beta=2.0$ & -- & -- & 7.86  & 11.05 & -0.11 & -0.40 \\
    $\beta=3.0$ & -- & -- & 8.59  & 12.34 & +0.61 & +0.89 \\
    \midrule
    \multicolumn{7}{l}{\textit{Effectiveness of Occlusion-Adaptive Voting}} \\
    \midrule
    Fixed threshold & -- & -- & 9.88 & 14.19 & +1.91 & +2.75 \\
    \textbf{Adaptive (ours)} & \textbf{--} & \textbf{--} & \textbf{7.97} & \textbf{11.44} & \textbf{--} & \textbf{--} \\
    \bottomrule
  \end{tabular}
  }
  \label{tab:ablation_robustness}
\end{table}
\section{Conclusion}

We investigate depth information for video individual counting in crowded scenes. Unlike RGB-based methods that mainly rely on temporal deduplication, our approach introduces depth constraints for per-frame detection and video-level inference. Depth-guided fusion and occlusion-aware prediction reduce missed detections and duplicate counting, while stable depth cues improve cross-frame association and video-level counting accuracy.

\balance
\bibliographystyle{ACM-Reference-Format}
\input{main.bbl}

\clearpage

\appendix
\section*{Appendix}

\section{RGBD-VideoCount Dataset Details}

RGBD-VideoCount is designed to mimic more realistic scenarios, where a single video typically contains multiple co-occurring categories, leading to challenging, mixed, and dense scenes. The dataset covers six everyday categories.

Beverage, Box, and Bag. These categories often appear in the same physical space and may exhibit similar appearance or containment relations. The beverage category includes three canonical scenarios: real drinks densely packed on shelves (high density and heavy occlusion), miniature beverages (medium occlusion), and vending machines (covering both severely occluded and almost unoccluded views). Multiple categories densely mixed in a limited space not only lead to severe occlusions but also impose higher demands on the model’s semantic discrimination ability.

Book and Cabinet. These categories are used to evaluate the model in highly regular and densely arranged scenes. Book videos depict multi-layer bookshelves packed with books; cabinet videos feature grid-like cabinet structures and, in some videos, dynamic door opening and closing, which place stringent requirements on localisation accuracy and fine-grained counting.

plant. This category provides diverse shapes and textures to assess the model’s generalisation ability further.

\section{Occlusion Score Construction.}
To characterize mutual occlusion among object instances in dense scenes, we define a continuous occlusion score for each instance, rather than using a simple binary occlusion label. Let $B_i$ denote the bounding box of instance $i$, and let $d_i$ denote the average depth value within $B_i$. For any other instance $j \neq i$, if it spatially overlaps with instance $i$ and is closer to the camera in depth, it is considered a potential occluder of instance $i$.

We first define the spatial occlusion strength of instance $j$ on instance $i$ as
\begin{equation}
s_{ij}=\frac{|B_i\cap B_j|}{|B_i|},
\label{eq:spatial_occ}
\end{equation}
where $|B_i\cap B_j|$ denotes the intersection area between the two bounding boxes, and $|B_i|$ denotes the area of instance $i$. This term measures the fraction of instance $i$ covered by instance $j$.

We then define a depth-ordering weight based on their depth difference:
\begin{equation}
z_{ij}=\sigma\left(\frac{d_i-d_j-\delta_z}{\tau_z}\right),
\label{eq:depth_weight}
\end{equation}
where $\sigma(\cdot)$ is the Sigmoid function, $\delta_z$ denotes the minimum effective depth gap, and $\tau_z$ is a smoothing parameter. When instance $j$ has a smaller depth value, i.e., it is closer to the camera, $z_{ij}$ becomes larger, indicating that it is more likely to occlude instance $i$.

Based on the above two terms, the occlusion contribution of instance $j$ to instance $i$ is defined as
\begin{equation}
c_{ij}=s_{ij}z_{ij}.
\label{eq:occ_contrib}
\end{equation}

Optionally, to suppress interference from weakly related overlaps, $c_{ij}$ can be further multiplied by a smooth IoU-based gating term. Finally, we aggregate the contributions of all neighboring instances to obtain the final occlusion score of instance $i$:
\begin{equation}
o_i=1-\prod_{j\neq i}(1-c_{ij}),
\label{eq:occ_score}
\end{equation}
where $o_i\in[0,1]$. A larger $o_i$ indicates that instance $i$ is more heavily occluded by foreground objects, while a smaller $o_i$ indicates that it is largely unoccluded.

During training, we use the continuous occlusion score $o_i$ as the supervision signal for the occlusion prediction branch, which learns the occlusion probability $p_{\mathrm{occ},i}$ for each candidate instance. During inference, the predicted occlusion probability is not only used to characterize single-frame detection results, but also further participates in subsequent trajectory-level occlusion statistics and temporal de-duplication.

\section{Probabilistic Interpretation of the Association Cost}
\subsection{Modality-Specific Likelihood Models}

For a trajectory $tr_i$ and a detection $det_j$, let the matching event be denoted by $M_{ij}=1$. We define the center displacement, center-depth residual, and normalized appearance features as
\begin{equation}
\Delta \mathbf{p}_{ij}=\mathbf{c}_i-\mathbf{c}_j,\qquad
\Delta z_{ij}=z_i-z_j,\qquad
\mathbf{u}_i,\mathbf{u}_j\in \mathbb{S}^{d-1}.
\end{equation}

\textbf{Positional Modality.}Under the online association setting with short temporal intervals, we adopt a local motion continuity assumption, i.e., a true match is more likely to appear near the previous trajectory location. Therefore, positional consistency is mainly characterized by the displacement magnitude rather than the specific motion direction. Based on this property, we model the positional modality as a radial decaying distribution over the displacement norm:
\begin{equation}
p(\Delta \mathbf{p}_{ij}\mid M_{ij}=1)\propto
\exp\!\left(-\frac{\|\Delta \mathbf{p}_{ij}\|_2}{\sigma_p}\right),
\end{equation}
where $\sigma_p$ denotes the effective noise scale of the positional modality. This form depends only on $\|\Delta \mathbf{p}_{ij}\|_2$ and is therefore isotropic.

\textbf{Depth Modality.}For the depth modality, we consider the one-dimensional residual consistency between a trajectory and a detection. In trajectory association, depth information is reflected by the temporal continuity of the target center depth across frames, so the absolute depth residual $|\Delta z_{ij}|$ serves as a natural observation. Meanwhile, depth estimation in real videos may be affected by occlusion, boundary ambiguity, and local distortions, causing the error distribution to typically exhibit a sharp peak with moderately heavy tails, i.e., small residuals dominate while occasional larger deviations may occur. Based on this characteristic, we adopt a Laplace residual model:
\begin{equation}
p(\Delta z_{ij}\mid M_{ij}=1)\propto
\exp\!\left(-\frac{|\Delta z_{ij}|}{\sigma_d}\right),
\end{equation}
where $\sigma_d$ denotes the effective noise scale of the depth modality.

\textbf{Appearance Modality.}For the appearance modality, we use $\ell_2$-normalized feature representations. As a result, the underlying geometry is no longer primarily characterized by Euclidean magnitude differences, but rather by directional consistency in the feature space. In this case, a more appropriate similarity measure is the angular relation on the unit hypersphere, namely the inner product or cosine similarity between normalized features. Accordingly, we write the appearance matching likelihood as
\begin{equation}
p(\mathbf{u}_j\mid \mathbf{u}_i,M_{ij}=1)\propto
\exp\!\left(\kappa_f\,\mathbf{u}_i^\top \mathbf{u}_j\right),
\end{equation}
where $\kappa_f$ denotes the concentration of appearance consistency. Since $\mathbf{u}_i^\top \mathbf{u}_j$ is equivalent to cosine similarity, the above form directly captures the relation that the more consistent the feature directions are, the higher the matching probability becomes.

\subsection{From the Matching Posterior to the Final Association Cost}

Given the matching event $M_{ij}=1$, we assume that the three modalities are
approximately conditionally independent, and introduce an IoU-based geometric prior
$p_{\mathrm{geo}}(M_{ij}=1)$. The matching posterior is written as
\begin{equation}
\begin{aligned}
p(M_{ij}=1 \mid tr_i, det_j)
\propto\;&
p(\Delta \mathbf{p}_{ij}\mid M_{ij}=1)\,
p(\Delta z_{ij}\mid M_{ij}=1) \\
&\cdot\,
p(\mathbf{u}_j\mid \mathbf{u}_i, M_{ij}=1)\,
p_{\mathrm{geo}}(M_{ij}=1).
\end{aligned}
\end{equation}

Substituting the likelihoods of the three modalities gives
\begin{equation}
\begin{aligned}
p(M_{ij}=1 \mid tr_i, det_j)
\propto\;&
\exp\!\left(-\frac{\|\Delta \mathbf{p}_{ij}\|_2}{\sigma_p}\right)
\exp\!\left(-\frac{|\Delta z_{ij}|}{\sigma_d}\right) \\
&\cdot\,
\exp\!\left(\kappa_f\,\mathbf{u}_i^\top \mathbf{u}_j\right)\,
p_{\mathrm{geo}}(M_{ij}=1).
\end{aligned}
\end{equation}

Taking the negative logarithm yields
\begin{equation}
\begin{aligned}
-\log p(M_{ij}=1 \mid tr_i, det_j)
=\;&
\frac{\|\Delta \mathbf{p}_{ij}\|_2}{\sigma_p}
+\frac{|\Delta z_{ij}|}{\sigma_d} \\
&-\kappa_f\,\mathbf{u}_i^\top \mathbf{u}_j
-\log p_{\mathrm{geo}}(M_{ij}=1)
+\mathrm{const}.
\end{aligned}
\end{equation}

Next, we rewrite the above terms using the distance forms adopted in the main text:
\begin{equation}
\begin{aligned}
d_{\mathrm{pos}}(tr_i,det_j)   &\coloneqq \|\Delta \mathbf{p}_{ij}\|_2, \\
d_{\mathrm{depth}}(tr_i,det_j) &\coloneqq |\Delta z_{ij}|, \\
d_{\mathrm{feat}}(tr_i,det_j)  &\coloneqq 1-\mathbf{u}_i^\top \mathbf{u}_j .
\end{aligned}
\end{equation}

Then the appearance term can be rewritten as
\begin{equation}
\begin{aligned}
-\kappa_f\,\mathbf{u}_i^\top \mathbf{u}_j
&=
\kappa_f\bigl(1-\mathbf{u}_i^\top \mathbf{u}_j\bigr)-\kappa_f \\
&=
\kappa_f\,d_{\mathrm{feat}}(tr_i,det_j)+\mathrm{const}.
\end{aligned}
\end{equation}

Therefore, the negative log-posterior becomes
\begin{equation}
\begin{aligned}
-\log p(M_{ij}=1 \mid tr_i, det_j)
=\;&
\frac{1}{\sigma_p}\,d_{\mathrm{pos}}^{ij}
+\kappa_f\,d_{\mathrm{feat}}^{ij}
+\frac{1}{\sigma_d}\,d_{\mathrm{depth}}^{ij} \\
&-\log p_{\mathrm{geo}}(M_{ij}=1)
+\mathrm{const},
\end{aligned}
\end{equation}
where
\[
d_{\mathrm{pos}}^{ij}\equiv d_{\mathrm{pos}}(tr_i,det_j),\quad
d_{\mathrm{feat}}^{ij}\equiv d_{\mathrm{feat}}(tr_i,det_j),\quad
d_{\mathrm{depth}}^{ij}\equiv d_{\mathrm{depth}}(tr_i,det_j).
\]

For the geometric prior, we adopt a hard IoU-gating constraint:
\begin{equation}
-\log p_{\mathrm{geo}}(M_{ij}=1)
=
\lambda_{\mathrm{iou}}\,
\mathbb{I}\!\bigl(\mathrm{IoU}(b_i,b_j)<\tau_{\mathrm{iou}}\bigr).
\end{equation}

As a result, the final association cost is
\begin{equation}
\begin{aligned}
C_{ij}
=\;&
w_p\,d_{\mathrm{pos}}(tr_i,det_j)
+w_f\,d_{\mathrm{feat}}(tr_i,det_j) \\
&+
w_d\,d_{\mathrm{depth}}(tr_i,det_j)
+\lambda_{\mathrm{iou}}\,
\mathbb{I}\!\bigl(\mathrm{IoU}(b_i,b_j)<\tau_{\mathrm{iou}}\bigr)
+\mathrm{const},
\end{aligned}
\end{equation}
with
\begin{equation}
w_p \propto \sigma_p^{-1},\qquad
w_d \propto \sigma_d^{-1},\qquad
w_f \propto \kappa_f.
\end{equation}

\section{Experiments on RGBD-VideoCount}
\subsection{Supplementary experiments}
We further supplement our comparisons by adopting the depth-enhanced detector YOLOv8 RGB-D as an additional baseline. Specifically, YOLOv8 RGB-D is used as the detection front-end and integrated with the same downstream tracking and counting pipeline as in the main experiments, with evaluation conducted under the same data split and metrics. The results(see Tab~\ref{tab:yolov8_rgbd_comparison}) show that, after introducing depth information at the detection stage, the performance is slightly better than that obtained with the RGB detector, indicating that depth cues can provide auxiliary benefits for target localization in densely arranged and occluded scenes. Nevertheless, our proposed model still achieves the best overall performance. This suggests that improvements in video instance counting do not solely rely on expanding the detector input modality, but also depend on the effective discrimination of adjacent similar instances, as well as accurate cross-frame association and deduplication.

\subsection{Additional Qualitative Results on RGBD-VideoCount}

\begin{figure*}[ht]
    \centering
    \includegraphics[width=\linewidth]{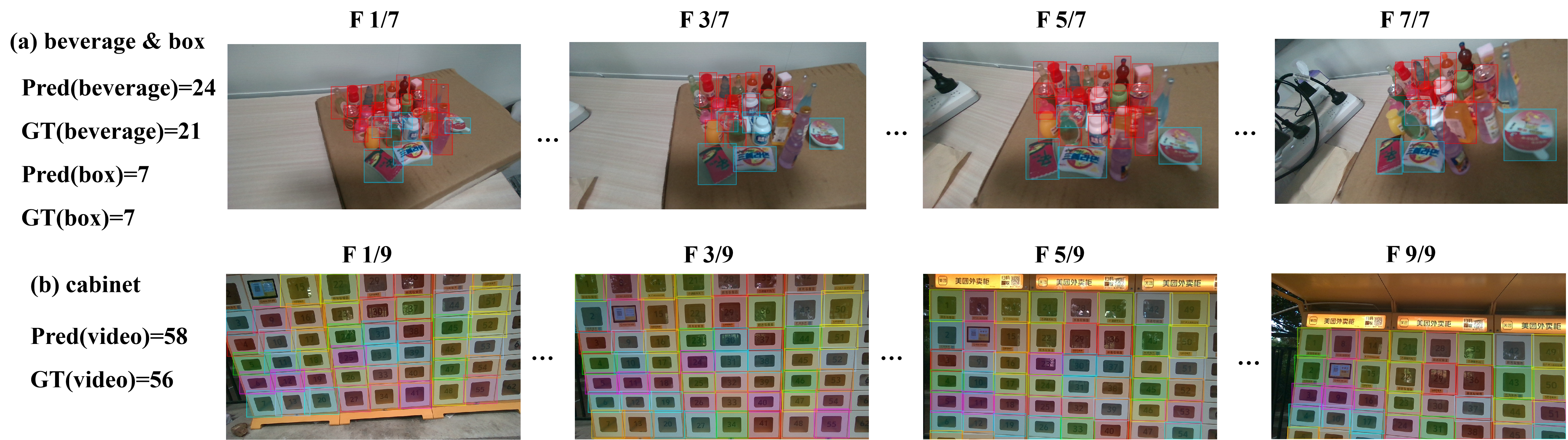}
    \caption{(a) Beverage scene: Demonstrates the model’s performance in occluded scenarios. (b) Cabinet scene: Demonstrates the model’s performance in dense scenes with highly homogeneous textures.}
    \label{fig:1}
\end{figure*}
\begin{table*}[t]
\centering
\caption{Comparison with YOLOv8 RGB-D based baselines on RGBD-VideoCount.}
\label{tab:yolov8_rgbd_comparison}
\resizebox{\textwidth}{!}{
\begin{tabular}{lcccccccccc}
\toprule
\multirow{2}{*}{Method} & \multirow{2}{*}{Detector} & \multirow{2}{*}{Tracker} 
& \multicolumn{4}{c}{RGBD-VideoCount Val} 
& \multicolumn{4}{c}{RGBD-VideoCount Test} \\
\cmidrule(lr){4-7} \cmidrule(lr){8-11}
& & 
& MAE$\downarrow$ & RMSE$\downarrow$ & AP$\uparrow$ & AP50$\uparrow$
& MAE$\downarrow$ & RMSE$\downarrow$ & AP$\uparrow$ & AP50$\uparrow$ \\
\midrule
YOLOv8 RGB-D & YOLOv8 RGB-D & SparseTrack 
& 25.36 & 42.85 & 16.42 & 31.95 
& 23.75 & 37.74 & 13.54 & 30.89 \\
YOLOv8 RGB-D & YOLOv8 RGB-D & SAM2.1 
& 20.32 & 41.81 & 17.42 & 43.41 
& 22.76 & 35.90 & 16.78 & 40.95 \\
\midrule
Ours & DG-Det & DG-Track 
& \textbf{13.32} & \textbf{19.68} & \textbf{23.96} & \textbf{50.12} 
& \textbf{7.97} & \textbf{11.44} & \textbf{20.78} & \textbf{50.39} \\
\bottomrule
\end{tabular}}
\end{table*}

\begin{table*}[ht]
\centering
\normalsize
\setlength{\tabcolsep}{5pt}
\renewcommand{\arraystretch}{1.10}

\caption{\textbf{Results on FSCD-147.}
\textcolor{bestred}{Best} results are marked in the reference-only settings, while
\textcolor{bestred}{best} and \textcolor{secondblue}{second-best} results are marked in the unified \emph{both} setting.}
\label{tab:fscd147_prompt_split}

\begin{tabular}{llcccccccc}
\toprule
\multirow{2}{*}{Prompt} & \multirow{2}{*}{Method} & \multicolumn{4}{c}{Val} & \multicolumn{4}{c}{Test} \\
\cmidrule(lr){3-6}
\cmidrule(lr){7-10}
& & MAE$\downarrow$ & RMSE$\downarrow$ & AP$\uparrow$ & AP50$\uparrow$
& MAE$\downarrow$ & RMSE$\downarrow$ & AP$\uparrow$ & AP50$\uparrow$ \\
\midrule
\multirow{6}{*}{text}
& Grounding DINO                 & 54.45 & 137.12 &  6.66 & 10.26 & 54.16 & 157.87 & 11.60 & 17.80 \\
& OWLv2                          & 49.02 & 131.41 & 11.39 & 20.33 & 41.83 & 149.82 & 22.84 & 35.76 \\
& PSeCo                          & 23.90 & 100.30 & $\times$ & $\times$ & 16.58 & 129.77 & \textcolor{bestred}{41.14} & \textcolor{bestred}{69.03} \\
& DAVE$_{\mathrm{prm}}$          & 14.87 & \textcolor{bestred}{59.60} & 16.31 & 47.12 & 15.52 & 114.10 & 18.50 & 50.24 \\
& \textsc{CountGD}               & 12.46 & 67.52 & 13.36 & 43.90 & 15.19 & 119.40 & 18.10 & 52.90 \\
& \textsc{CountGD-Box}           & \textcolor{bestred}{12.24} & 66.24 & \textcolor{bestred}{27.02} & \textcolor{bestred}{60.73} & \textcolor{bestred}{15.01} & \textcolor{bestred}{118.16} & 30.44 & 61.56 \\
\midrule
\multirow{6}{*}{exemplar}
& Counting-DETR                  & 20.38 & 82.45 & 17.27 & 41.90 & 16.79 & 123.56 & 22.66 & 50.57 \\
& PSeCo                          & 15.31 & 58.34 & 32.12 & 60.02 & 13.05 & 112.86 & \textcolor{bestred}{43.53} & 74.64 \\
& DAVE                           &  9.75 & \textcolor{bestred}{40.30} & 24.20 & 61.08 & 10.45 & 74.51 & 26.81 & 62.82 \\
& GeCo                           & 9.52 & 43.00 & \textcolor{bestred}{33.51} & 62.51 & \textcolor{bestred}{7.91} & \textcolor{bestred}{54.28} & 43.42 & \textcolor{bestred}{75.06} \\
& \textsc{CountGD}               &  9.34 & 51.74 & 15.16 & 47.96 & 10.77 & 99.51 & 19.76 & 57.24 \\
& \textsc{CountGD-Box}           & \textcolor{bestred}{9.27} & 52.15 & 32.09 & \textcolor{bestred}{68.82} & 10.85 & 99.60 & 34.81 & 69.46 \\
\midrule
\multirow{3}{*}{both}
& \textsc{CountGD}               & \textcolor{bestred}{\textbf{8.69}} & \textcolor{bestred}{\textbf{43.89}} & 14.74 & \textcolor{secondblue}{47.95} & \textcolor{bestred}{\textbf{10.18}} & \textcolor{secondblue}{96.20} & 20.50 & \textcolor{secondblue}{59.40} \\
& \textsc{CountGD-Box}           & \textcolor{secondblue}{8.76} & \textcolor{secondblue}{44.24} & \textcolor{bestred}{33.10} & \textcolor{bestred}{\textbf{71.38}} & \textcolor{secondblue}{10.29} & 96.33 & \textcolor{bestred}{\textbf{36.20}} & \textcolor{bestred}{\textbf{72.39}} \\
& \textbf{Ours}                  & 14.85 & 51.86 & \textcolor{secondblue}{20.45} & 47.22 & 11.91 & \textcolor{bestred}{\textbf{93.78}} & \textcolor{secondblue}{26.58} & 55.95 \\
\bottomrule
\end{tabular}

\vspace{2mm}

\begin{minipage}{\textwidth}
  \centering
  \includegraphics[width=\textwidth,height=0.26\textheight,keepaspectratio]{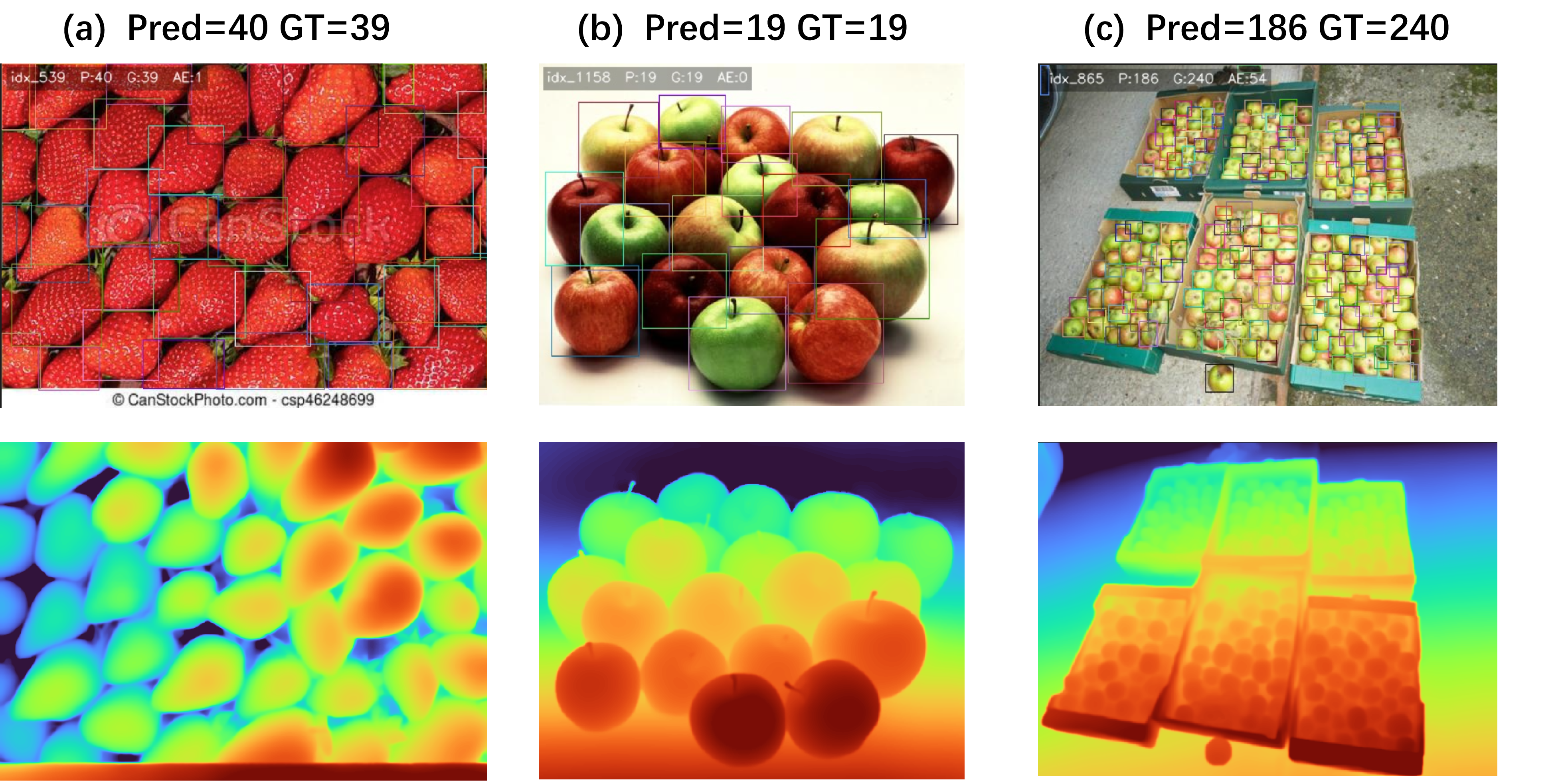}
  \captionof{figure}{\textbf{Qualitative examples on FSCD-147 with pseudo-depth.}
  Top row: input images with predicted counts (Pred) and ground-truth counts (GT).
  Bottom row: pseudo-depth maps generated by Depth Anything V2.
  In (a) and (b), the pseudo-depth provides relatively clear geometric cues and the predictions are accurate.
  In (c), the model yields a suboptimal prediction due to the influence of the pseudo-depth.}
  \label{fig:6}
\end{minipage}

\vspace{-2mm}
\end{table*}

We present two representative scene categories, \emph{Beverage \& Box} and \emph{Cabinet}, in Fig.~\ref{fig:1}.

\textbf{Beverage \& Box:} This scene is a simulated densely arranged setting involving beverages and boxes. Multiple objects are placed within a limited space, resulting in evident occlusions, overlaps, and viewpoint changes, while different objects also exhibit high similarity in appearance, size, and packaging form. These factors make the boundaries between adjacent objects more difficult to distinguish, often leading to missed detections, mismatches, and duplicate counting. Nevertheless, our method can still stably distinguish closely adjacent instances in this scenario, demonstrating strong capability in dense-scene perception and counting.
\textbf{Cabinet:} This category represents a typical dense setting, where cabinet compartments are tightly arranged with highly homogeneous textures. Our model exhibits strong fine-grained localization and can accurately discriminate individual storage units, demonstrating robust performance on densely packed and visually similar objects. 

\section{Experiments on FSD-147}
Our method enhances \textsc{CountGD-Box} by fusing an additional depth modality to better handle occlusions and scale variations. To evaluate its generalization, we extend our experiments to the FSCD-147 dataset and use pseudo-depth maps generated by Depth Anything V2 as a surrogate input. Due to the limited quality of the estimated depth, incorporating this modality does not yield the expected substantial gains. In particular, the pseudo-depth often falls short in preserving object-boundary details, matching the noise characteristics of real depth sensors, and maintaining consistent absolute scale, which makes it difficult for the model to extract stable and accurate geometric cues.

As shown in Fig.~\ref{fig:6}(c), while the pseudo-depth from Depth Anything V2 captures the coarse depth layout, it is over-smoothed and blurs depth discontinuities (e.g., box boundaries), failing to resolve instance-level depth differences in densely packed apples.  Consequently, the model cannot reliably extract depth cues from such pseudo-depth, thereby weakening the discriminative power of the depth modality in crowded scenes and limiting the effectiveness of the resulting predictions.

\section{Failure Scenes analysis}

\begin{figure*}[t]
    \centering
    \includegraphics[width=\textwidth]{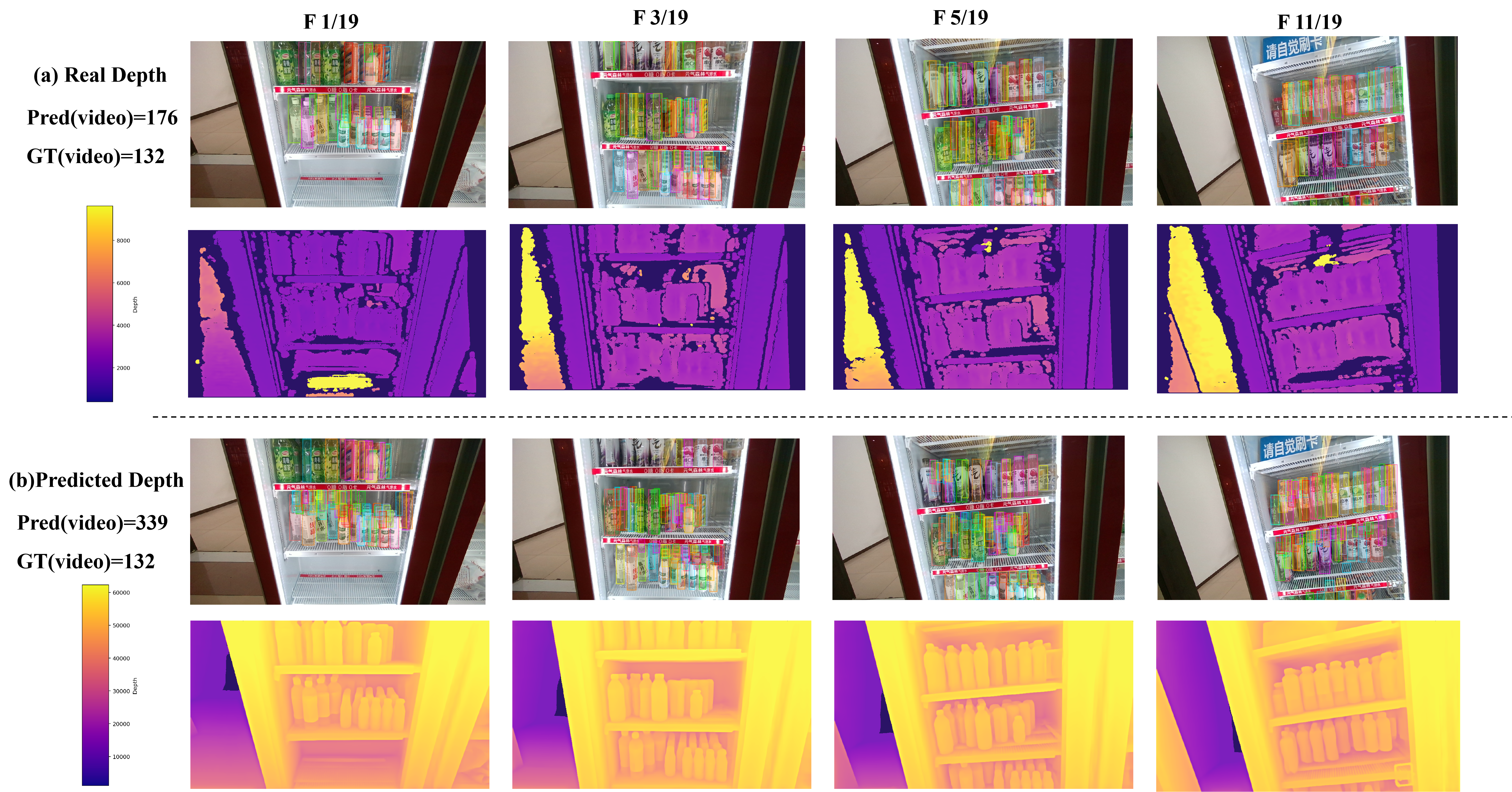}
    \caption{Failure cases in glass-door beverage cabinet scenes. The first and third rows show the RGB images overlaid with detection results, while the second and fourth rows present the corresponding raw depth maps and predicted depth maps. }
    \label{fig:failure_glass_depth}
\end{figure*}

As shown in the figure~\ref{fig:failure_glass_depth}, in beverage scenes with glass doors, the glass surface introduces reflections, refractions, and local highlights, which cause depth values acquired by the depth sensor to exhibit noise, missing regions, or structural distortions. This issue is particularly pronounced around the cabinet door boundaries, reflective glass regions, and densely arranged bottles and cans, where the raw depth map often fails to accurately represent the true spatial hierarchy of the objects, thereby affecting the effective separation between foreground targets and background structures. In such scenarios, if the model directly relies on raw depth information, it is easily disturbed by abnormal depth responses, which further leads to unstable target localization, ambiguous boundaries between adjacent instances, and even missed detections or duplicate counting. 

Meanwhile, the predicted depth information also performs poorly in this type of scene. Although the predicted depth maps can maintain relatively smooth global structures in some regions, the reflections from glass, transparent occlusions, and complex illumination conditions disrupt the stable correspondence between image appearance and true scene geometry. As a result, the model often fails to recover accurate depth hierarchies, and the predicted depth still suffers from structural deviations and blurred boundaries. Therefore, predicted depth cannot effectively compensate for the limitations of raw depth in glass scenes, and its contribution to dense instance separation and counting remains limited.

This failure case indicates that, in scenes involving glass, transparent media, and strong reflections, depth information itself may lose its expected geometric discriminative power. Neither directly captured raw depth nor model-estimated predicted depth can provide stable and reliable support for video instance counting. Therefore, improving the robustness of the model under abnormal depth inputs, as well as more effectively integrating RGB appearance cues with unreliable depth signals in transparent and reflective scenes, remains an important direction for future research.

\end{document}